%% file: main.tex
\documentclass[preprint,12pt,authoryear]{elsarticle}

\usepackage{amssymb}
\input{packages}

\input{acronyms}

\journal{International Journal of Applied Earth Observation and Geoinformation}

\begin{document}

\newcommand{\deepaqua}{\textsc{DeepAqua}\xspace}

\begin{frontmatter}

\title{\deepaqua: Self-Supervised Semantic Segmentation of Wetland Surface Water Extent with \acrshort{sar} Images using Knowledge Distillation}

\author[su,kth]{Francisco J. Pe\~na}

\author[su]{Clara H\"ubinger}

\author[kth]{Amir H. Payberah}

\author[su]{Fernando Jaramillo}

\affiliation[su]{organization={Department of Physical Geography, Stockholm University},%
            addressline={Svante Arrhenius v\"ag 8}, 
            city={Stockholm},
            postcode={106 91}, 
            country={Sweden}}

\affiliation[kth]{organization={Division of Software and Computer Systems, KTH Royal Institute of Technology},%
            addressline={Kistag{\aa}ngen 16}, 
            city={Kista},
            postcode={164 40}, 
            country={Sweden}}

\begin{abstract}
\input{0_abstract}

\end{abstract}


\end{frontmatter}

\input{1_introduction}
\input{2_related_work}
\input{3_background}
\input{4_methodology}
\input{5_evaluation}
\input{6_conclusion}

\input{acknowledgements}

\bibliographystyle{elsarticle-harv}
\bibliography{bibliography}

\end{document}

%% file: packages.tex
\usepackage[T1]{fontenc}
\usepackage[acronym]{glossaries}
\usepackage{float}
\usepackage{booktabs}
\usepackage[dvipsnames]{xcolor}
\usepackage{todonotes}
\usepackage{lineno}
\usepackage{enumitem}
\usepackage{graphicx}
\usepackage{array}
\usepackage{xspace}
\usepackage{amsmath,amssymb,amsfonts}
\usepackage{pgfplots}
\usepackage{subcaption}

%% file: acronyms.tex
\newacronym{awei}{AWEI}{Automated Water Extraction Index}
\newacronym[plural=CNNs,firstplural=Convolutional Neural Networks (CNNs)]{cnn}{CNN}{Convolutional Neural Network}
\newacronym{fcn}{FCN}{Fully Convolutional Network}
\newacronym{fp}{FP}{False Positives}
\newacronym{fn}{FN}{False Negatives}
\newacronym[plural=GANs,firstplural=Generative Adversarial Networks (GANs)]{gan}{GAN}{Generative Adversarial Network}
\newacronym{hrwi}{HRWI}{High Resolution Water Index}
\newacronym{iou}{IOU}{Intersection Over Union}
\newacronym{mndwi}{MNDWI}{Modified NDWI}
\newacronym{ndwi}{NDWI}{Normalized Difference Water Index}
\newacronym{nir}{NIR}{Near-Infrared}
\newacronym{pa}{PA}{Pixel Accuracy}
\newacronym{rnn}{RNN}{Recurrent Neural Network}
\newacronym{sar}{SAR}{Synthetic Aperture Radar}
\newacronym{tp}{TP}{True Negatives}
\newacronym{tn}{TN}{True Negatives}
\newacronym{tsuwi}{TSUWI}{Two-step Urban Water Index}
\newacronym{vgg}{VGG}{Visual Geometry Group}

%% file: 0_abstract.tex
Deep learning and remote sensing techniques have significantly advanced water monitoring abilities; however, the need for annotated data remains a challenge. This is particularly problematic in wetland detection, where water extent varies over time and space, demanding multiple annotations for the same area. In this paper, we present \deepaqua, a self-supervised deep learning model that leverages knowledge distillation (a.k.a. teacher-student model) to eliminate the need for manual annotations during the training phase. We utilize the \acrfull{ndwi} as a teacher model to train a \acrfull{cnn} for segmenting water from \acrfull{sar} images, and to train the student model, we exploit cases where optical- and radar-based water masks coincide, enabling the detection of both open and vegetated water surfaces. \deepaqua represents a significant advancement in computer vision techniques by effectively training semantic segmentation models without any manually annotated data. Experimental results show that \deepaqua outperforms other unsupervised methods by improving accuracy by 7\%, \acrlong{iou} by 27\%, and F1 score by 14\%. This approach offers a practical solution for monitoring wetland water extent changes without needing ground truth data, making it highly adaptable and scalable for wetland conservation efforts.

%% file: 1_introduction.tex
\section{Introduction}\label{sec:introduction}

Wetlands provide essential ecosystem services such as water purification, flood regulation, and carbon sequestration, and are critical for sustainable development \citep{jaramillo2019priorities}. They are, however, increasingly threatened by climate change and human activity \citep{thorslund2017wetlands}. The comprehensive monitoring of wetland surface water extent, including both open and vegetated water surfaces, is essential for the effective wetland monitoring needed for their conservation and management.

Advancements in remote sensing techniques have greatly improved surface water monitoring capabilities \citep{feyisa2014automated}. While optical sensors are adept at detecting open water surfaces, they struggle to identify vegetated waters, underestimating wetland water extent. Radar sensors, such as \gls{sar}, can overcome this limitation by penetrating the vegetation \citep{banks2019wetland}. Still, analyzing \gls{sar} images can be challenging due to noise and speckles, making it difficult to distinguish water from soil and vegetation. 

In recent years, deep learning techniques for the semantic segmentation of images have shown promise in addressing these challenges. However, the substantial amount of annotated data required for training these models is often time-consuming and costly, creating significant barriers to their widespread adoption. Furthermore, existing methods struggle to accurately identify vegetated waters, underestimating wetland surface water extent.

In this paper, we present \deepaqua, a self-supervised deep learning model that employs knowledge distillation (a.k.a teacher-student model) to eliminate the need for manual annotation during the training phase. Utilizing the \gls{ndwi}~\citep{mcfeeters1996use} as the ``teacher'' model, we train a ``student'' U-Net~\citep{ronneberger2015u} to recognize water boundaries in any \gls{sar} image. Using the signal from non-vegetated water as a guide, the student model learns from the teacher model, eventually recognizing both open and vegetated water bodies. The \gls{ndwi} ``teacher'' model can generate annotated images of water surfaces without training. Here, we address the problem of how to train a deep learning model to recognize surface water from radar imagery. We aim to eliminate the costs of collecting training data through fieldwork or manual annotations of radar images.

The contributions of our paper are as follows:

\begin{itemize}
    \item We present \deepaqua, a novel method for training semantic segmentation models using knowledge distillation without any manually annotated data.
    \item Our method employs \gls{ndwi} masks as proxies for semantic labels and optical images as an auxiliary modality to supervise a \gls{sar}-based U-Net.
    \item We present \deepaqua as a highly adaptable and scalable model, as it does not require ground truth for training.
    \item We enhance the efficiency of wetland conservation strategies, as \deepaqua can monitor changes in surface water coverage, encompassing both exposed and vegetated aquatic areas. 
\end{itemize}

%% file: 2_related_work.tex
\section{Related Work}\label{sec:related_work}
The challenge of accurately detecting wetlands — a critical task for environmental and societal applications such as flood mapping and water resource management — is intensified by variable spectral signatures of water resulting from illumination, turbidity, and vegetation. Multispectral optical imagery from satellites such as Sentinel-2 has been widely used to map wetlands using deep learning methods \citep{jiang2019arcticnet,cui2020wetlandnet,dang2020coastal,jamali2021wetland,pham2022new,onojeghuo2023wetlands}. For example, \citep{rezaee2018deep} fine-tunes an AlexNet \citep{krizhevsky2017imagenet} pre-trained on the ImageNet \citep{deng2009imagenet} dataset to classify wetland patches using optical imagery. \cite{mahdianpari2018very} explores multiple \gls{cnn} architectures to determine which architecture produced the most accurate wetland classifications. However, these approaches cannot detect water hidden under vegetation, which is crucial for monitoring water extent in wetlands.

Radar imagery from satellites such as Sentinel-1, which has a \gls{sar} C-band sensor that can penetrate vegetation and clouds \citep{geudtner2014sentinel}, have additionally been used for this purpose. \cite{slagter2020mapping} combined Sentinel-1 with optical Sentinel-2 and fieldwork data to map wetlands using a random forest classifier. The WetNet model \citep{hosseiny2021wetnet} is an ensemble of three classifiers that uses multitemporal images to map wetlands: a 2D-CNN trained on radar imagery, a 3D-CNN trained on multispectral and multitemporal imagery, and a \gls{rnn} trained with multivariate temporal information. \cite{jamali20223dunetgsformer} introduces the 3DUnetGSFormer model, which uses a \gls{gan} \citep{goodfellow2020generative} to generate synthetic data with similar characteristics as the ground-truth data and a Swin transformer \citep{liu2021swin} to classify the wetland images. Similar approaches have used optical and radar imagery to map wetlands, such as \cite{jamali2021deep} and \cite{jamali2022swin}.

All these approaches have a common limitation: \emph{they require manually annotated data to train their models}. The manually annotated data usually come from fieldwork, which is very costly and time-consuming to acquire due to logistics, equipment maintenance, sampling, etc. Moreover, these approaches assume that the surface water extent is constant over time, although the water extent usually varies across the season and is subject to weather conditions. For instance, one wetland location could have been labeled ``open water'' because the image was taken in April when the surface water extent of the wetland increased due to snowmelt. On the other hand, the same location could be dry in July, leading to inaccurate predictions and quantification of water surfaces.

%% file: 3_background.tex
\section{Background and Problem Formulation} \label{sec:background}

This paper addresses the problem of \emph{detecting water surfaces under vegetation without requiring fieldwork or manually annotated data}. To create a model and train it without fieldwork or manually annotated data, we combine remote sensing techniques (Section \ref{sec:remote_sensing}) with deep learning techniques (Sections \ref{sec:cnn}, \ref{sec:distillation}, and \ref{sec:self_supervised}). Here, we recall some of their basic concepts.

\subsection{Detecting Surface Water using Remote Sensing} \label{sec:remote_sensing}

\begin{figure}[H]
    \centering
    \includegraphics[width=\linewidth]{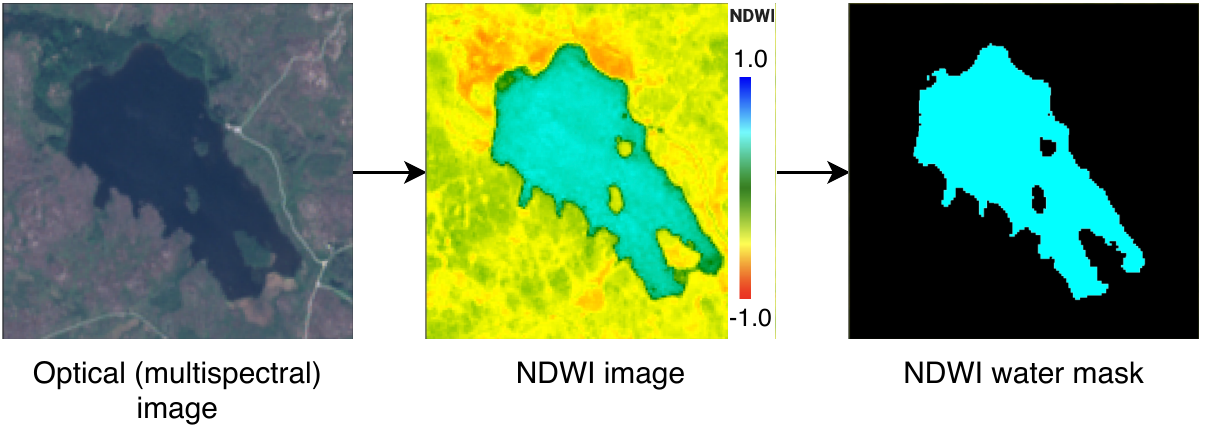}
    \caption{Water delineation using \acrshort{ndwi}. The left image comprises red, green, and blue bands (RGB). The middle image is generated using the \acrshort{ndwi} index. The right image shows an \acrshort{ndwi} image where pixels with positive values are cyan, and the rest are black, thus completing the water delineation process.}
    \label{fig:ndwi}
\end{figure}

Traditionally, optical sensors based on reflected solar radiation have been used to detect water. \gls{ndwi} is one of the most popular optical methods to delineate open waters \citep{mcfeeters1996use}, as it helps differentiate open water from soil since: (1) water reflects green light, (2) water has low reflectance of \gls{nir} light, and (3) terrestrial vegetation and soil have a high reflectance of \gls{nir} light. For each pixel in an image, the \gls{ndwi} is calculated using the normalized difference between the green light intensity and the \gls{nir} light intensity -- the results of the \gls{ndwi} index range from -1 to +1. Water surfaces have positive values, while soil and terrestrial vegetation have zero or negative values because they typically have a higher reflectance of \gls{nir} than green light. Figure \ref{fig:ndwi} shows how \gls{ndwi} is used to delineate open water.

Other index-based methods to detect water from optical imagery include the \gls{mndwi} \citep{xu2006modification}, \gls{hrwi} \citep{yao2015high}, \gls{tsuwi} \citep{wu2018two} and \gls{awei} \citep{feyisa2014automated}.

Although \gls{ndwi} provides reliable information on open water, its usage is limited to cloud-free days. Areas with low albedo and shadows can introduce inaccuracies in detection \citep{feyisa2014automated}. On the other hand, \gls{sar} does not depend on sunlight and is able to penetrate vegetation, presenting an attractive alternative to optical imagery \citep{mondini2021landslide}. As such, \gls{sar} can provide data on water hidden under vegetation. However, \gls{sar} sensors provide complex information that can be challenging to interpret. Moreover, water may be mistaken with unvegetated or sparsely vegetated areas, adding additional difficulties to the automatic detection of water in \gls{sar} images \citep{tsyganskaya2018detection,hardy2019automatic}.

\subsection{Semantic Segmentation of Images} \label{sec:cnn}

Semantic segmentation is a computer vision technique to identify and classify objects within an image at the pixel level. Instead of just identifying objects as a whole, semantic segmentation identifies the exact boundaries of each object and assigns a specific label or class to each pixel within the object's boundary. This allows for a more precise and detailed analysis of an image, making it useful for applications such as self-driving cars, medical imaging, and, in our case, water delineation.

\glspl{cnn} are one of the most efficient deep learning approaches for image processing \citep{lecun2015deep}. They work by reading an image through a series of layers that extract features such as edges and shapes and then using those features to recognize patterns in the image. They extract a varying level of abstraction from the data in different layers with the added benefit of not requiring prior feature extraction and having more generalization capability. \glspl{cnn} use convolutional layers to extract features, pooling layers to downsample the output, and activation functions to introduce non-linearity. \glspl{cnn} have proven effective in image recognition tasks because they can learn to recognize patterns and features in images without being explicitly programmed. \glspl{cnn} have better prediction accuracy because they (1) can retain the geometrical properties from two-dimensional images, (2) can be trained with large amounts of data and perform consistently across varied data, and (3) do not require expert input of features; the \glspl{cnn} can learn the features.

We use a particular \gls{cnn} architecture called U-Net \citep{ronneberger2015u} for producing semantic segmentations. U-Net is designed in a way that it has a contracting path followed by an expanding path. The contracting path consists of a series of convolutional layers, reducing the spatial dimensions of the image while increasing the number of channels. This helps to extract high-level features from the input image. The expanding path then consists of a series of up-convolutional layers, which upsample the feature maps to restore the original spatial dimensions of the image, generating a segmentation map with the same dimensions as the input image. Other architectures for semantic segmentation include Resnet \citep{he2016deep}, MSResNet \citep{dang2021msresnet}, and MSCENet \citep{kang2021multi}. 

Here, we use knowledge distillation to train a U-Net without requiring manually annotated data. 

\subsection{Knowledge Distillation} \label{sec:distillation}

Knowledge distillation \citep{hinton2015distilling,xu2020knowledge}, which is also known as the teacher-student model, is a process of teaching a smaller and simpler model - the student model - to mimic the behavior of a larger and more complex model - the teacher model - to achieve similar performance. During the distillation process, the teacher model generates predictions for training data, which are then used to train the student model. However, instead of training the student model to directly predict the correct output, the student model is trained to learn from the teacher model's predictions.

By doing so, the student model can learn from the teacher model, including the relationships between different input features and the patterns in the data, which is difficult to learn from the training data alone. The result is a smaller and faster model that performs similarly to the larger and more complex teacher model. Additionally, the student model may generalize better than the teacher model in specific scenarios \citep{deng2022personalized,beyer2022knowledge}, as it has learned to capture the most important aspects of the teacher's behavior while ignoring the noise. In this paper, we tweak the knowledge distillation process: instead of having a small model learn from a large model, we make a {\em radar-based} model learn from an {\em optical-based} model. We exploit the fact that it is easier to identify water from optical images rather than from radar images. This process is called cross-modal knowledge distillation \citep{hu2020creating}. By using knowledge distillation, we create a self-supervised model. 

\subsection{Self-Supervised Learning} \label{sec:self_supervised}

One of the biggest bottlenecks in deep learning models, including \glspl{cnn}, is requiring large amounts of annotated data to make accurate predictions. Particularly in semantic segmentation of images, manually annotating each image is costly and time-consuming, aggravated by the fact that deep learning models require thousands of images to produce accurate predictions. Techniques like transfer learning \citep{garcia2018survey} and data augmentation \citep{shorten2019survey} have alleviated the need for large amounts of annotated data. Nevertheless, they still require a minimum amount of data. Self-supervised learning is a machine learning method that allows algorithms to learn without needing human-annotated samples \citep{shurrab2022self}.

The following section shows how we achieve self-supervision using knowledge distillation architecture. We automatically generate ground truth data using \gls{ndwi} and train a \gls{sar}-based \gls{cnn} to detect water. Our approach has the advantage of requiring \emph{\textbf{zero}} manually annotated data.

%% file: 4_methodology.tex
\section{\deepaqua Model}\label{sec:methodology}

\begin{figure}[H]
  \centering
  \begin{subfigure}[b]{\linewidth}
    \includegraphics[width=\linewidth]{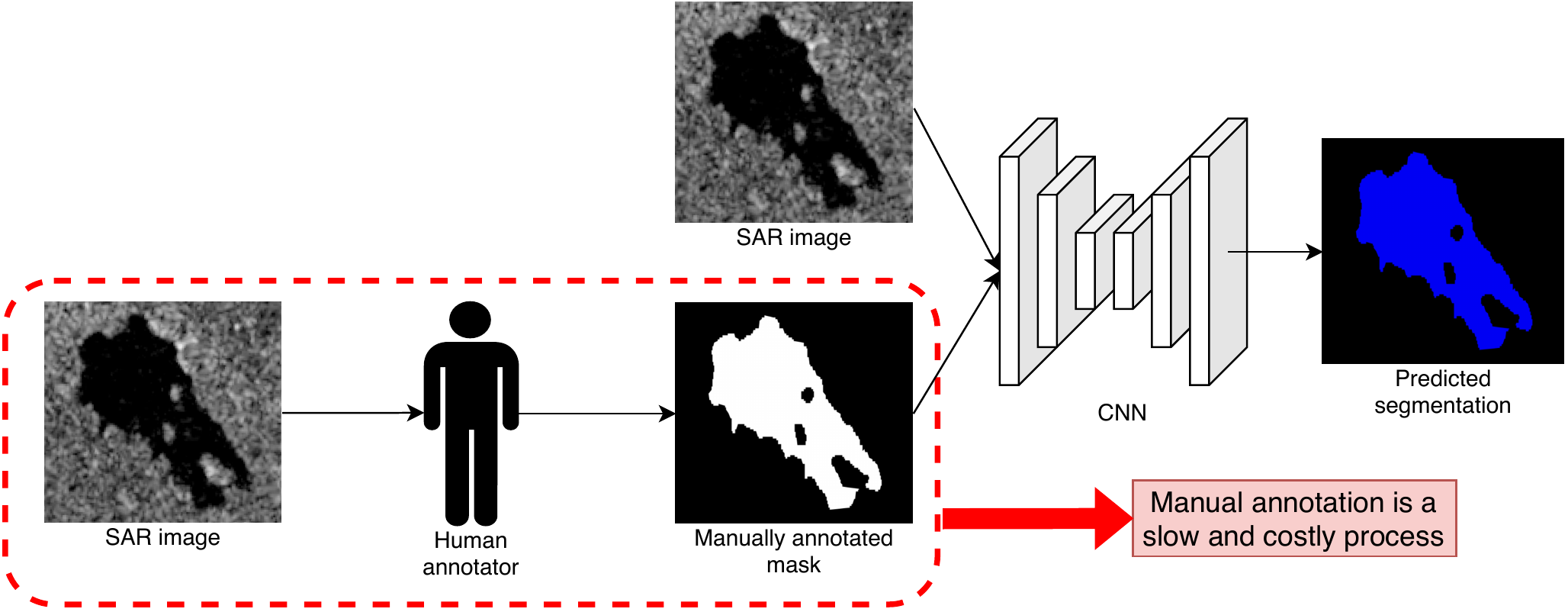}
    \caption{The traditional methodology for training deep learning models relies on manually annotated data.}
    \label{fig:traditional_training}
  \end{subfigure}
  
  \vspace{1em} 
  
  \begin{subfigure}[b]{\linewidth} 
    \includegraphics[width=\linewidth]{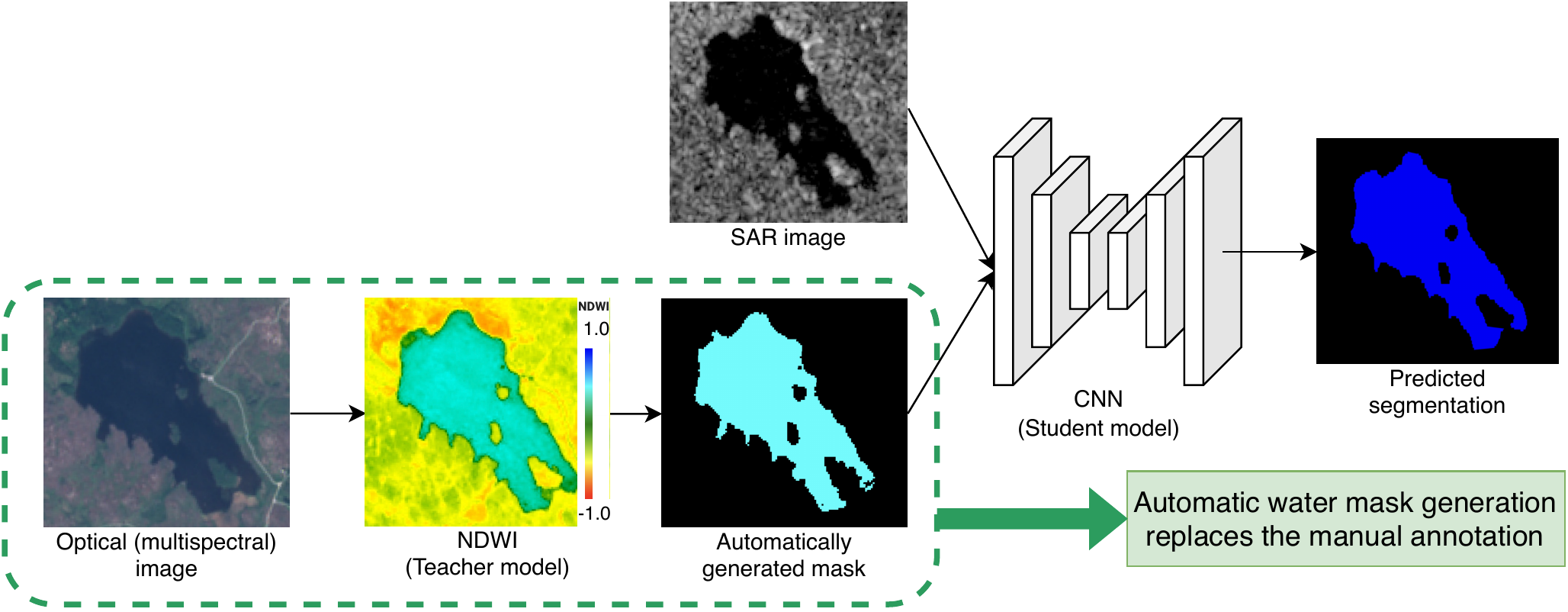}
    \caption{Our proposed methodology automatizes the data annotation process.}
    \label{fig:self_supervised_training}
  \end{subfigure}
  
  \caption{The training of a \acrshort{cnn} model to recognize water boundaries from \acrshort{sar} imagery using \acrshort{ndwi} water masks as ground truth.}
  \label{fig:training_architectures}
\end{figure}

Semantic segmentation projects typically involve a human annotator delineating images to indicate which parts correspond to a particular object or feature, such as water surfaces in radar images. This process can be time-consuming and expensive, often making data annotation the bottleneck of deep learning projects. Figure \ref{fig:traditional_training} shows this traditional architecture with a human annotator delineating the water in radar images and a \gls{cnn} trained based on these annotated images to recognize water in previously unseen images.

In this study, as seen in Figure \ref{fig:self_supervised_training}, instead of relying on human-annotated water masks, our knowledge distillation architecture uses the \gls{ndwi} model as the teacher and the \gls{cnn} as the student, where the teacher extracts knowledge about the location of the water surface from optical imagery and produces segmented images that the student will try to mimic. Unlike most knowledge distillation approaches, our teacher and student models rely on different data types; particularly, the teacher uses optical imagery to produce water segmentations, while the student uses radar imagery. For each batch of images, we calculate the Dice loss \citep{soomro2018strided} between the student predictions and the ground truth masks provided by the teacher and minimize this loss function to improve the performance of the student model. We then backpropagate the loss to update the weights of the \gls{cnn}. By eliminating the need for manual annotation, we aim to streamline the model training process and reduce overall project costs.

\subsection{\deepaqua Framework and Workflow}

\begin{figure}[H]
    \centering
    \includegraphics[width=\linewidth]{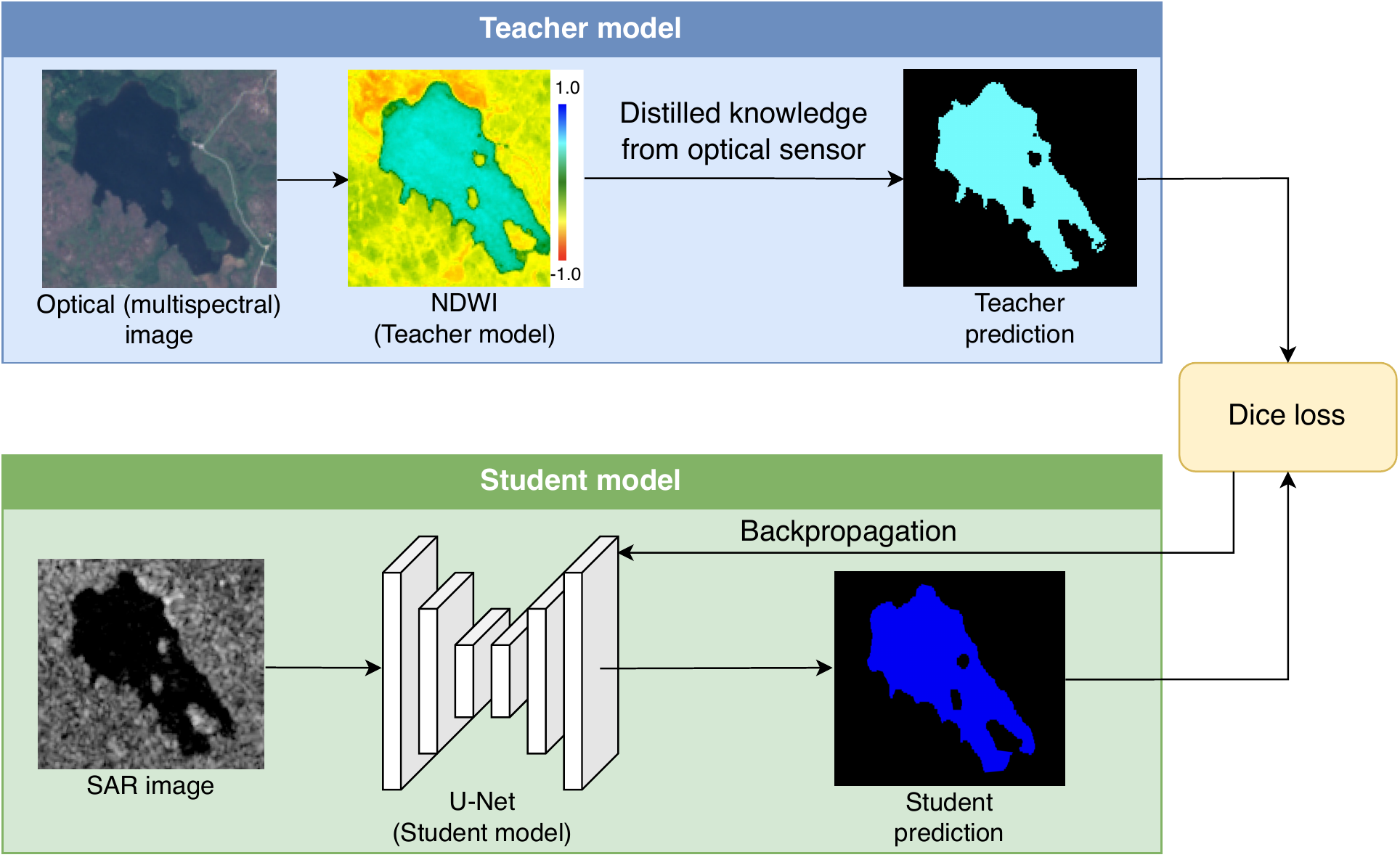}
    \caption{The process of training a \acrshort{cnn} model to recognize water boundaries from \acrshort{sar} imagery by learning from a \acrshort{ndwi} model.}
    \label{fig:entire_pipeline_dice_loss}
\end{figure}

Figure \ref{fig:entire_pipeline_dice_loss} shows the \deepaqua's overall framework and workflow. Our method consists of two models: teacher and student models. The teacher model is a thresholding model that generates water masks by applying the \gls{ndwi} index to optical images, and the student model which is a U-Net \citep{ronneberger2015u} that takes \gls{sar} images as input and produces segmentation masks as output. The teacher and student models are trained jointly by minimizing the Dice loss between their outputs. The workflow of our method is as follows:

\begin{itemize}
    \item Step 1: We create a training set by selecting images that fulfil the following conditions: (1) Sentinel-1 and Sentinel-2 image availability on the same date for the region of interest, (2) a maximum of 1\% of cloud cover on the Sentinel-2 image, and (3) no missing values on the Sentinel-1 and Sentinel-2 images.
    \item Step 2: Given a pair of optical and \gls{sar} images that are co-registered and cover the same geographic area, we feed the optical image to the teacher model and obtain an \gls{ndwi} mask as its output.
    \item Step 3: We feed the \gls{sar} image to the student model and obtain a segmentation mask as its output.
    \item Step 4: We compute the Dice loss between the teacher and student output to measure their similarity.
    \item Step 5: We update the student weights using backpropagation based on the Dice loss.
    \item Step 6: We repeat steps 2-5 for all pairs of optical and \gls{sar} images in the training set until convergence.
\end{itemize}

\subsection{The Teacher Model}

The teacher model generates water masks from optical images using the \gls{ndwi} index. Selecting the optimal threshold for \gls{ndwi} values is challenging, as highlighted by \cite{ji2009analysis} and \cite{reis2021uncertainties}. Following the recommendation of \cite{mcfeeters1996use}, we adopted a threshold value of 0.0 to delineate open water. The teacher model outputs a binary water mask matching the input optical image size, where 0 indicates ground and 1 denotes water. This mask then guides the student model.

\subsection{The Student Model}

The student model is a U-Net \citep{ronneberger2015u} model that takes \gls{sar} images as input and produces segmentation masks as output. We use U-Net as the student model for two reasons. First, U-Net is a simple and effective model for semantic segmentation that can achieve good results with limited data and computational resources. Second, U-Net is compatible with the teacher model regarding input and output sizes, facilitating the cross-modal knowledge distillation process.

We train U-Net from scratch on \gls{sar} images without requiring annotated data. We use \gls{sar} images from Sentinel-1, a satellite mission that provides C-band \gls{sar} images with a resolution of $10$m $\times 10$m per pixel. For enhanced contrast in these images, we exclude pixel values below the 1st percentile and above the 99th percentile. Subsequently, we normalize the images. The output of the U-Net is a segmentation mask that has the same size as the input \gls{sar} image. Within this mask, values span from 0 to 1, with higher values suggesting increased water presence. This segmentation mask serves as the desired output for model optimization.

\subsection{The Cross-Modal Knowledge Distillation Process} \label{sec:loss}

The cross-modal knowledge distillation process is the core of our method that transfers knowledge from the teacher to the student model. The teacher model produces a hard \gls{ndwi} water mask from an optical image, and the student model produces a segmentation mask from a \gls{sar} image. The hard \gls{ndwi} mask and the segmentation mask are aligned in terms of spatial resolution and geographic area, as they are generated from co-registered optical and \gls{sar} images that cover the same scene. The cross-modal knowledge distillation process aims to minimize the Dice loss between the \gls{ndwi} mask and the segmentation mask, which measures their similarity.

The Dice loss \citep{dice1945measures} is a loss function frequently employed for semantic segmentation tasks. It's particularly suited for imbalanced data, where one class (e.g., water pixels) might be significantly underrepresented compared to another (e.g., land pixels). The Dice loss is defined as:

\begin{equation}
    \label{eq:dice_loss}
    \mathcal{L}_{Dice} = 1 - \frac{2 \times |Y_{T} * Y_{S}| + \epsilon}{|Y_{T}| + |Y_{S}| + \epsilon}
\end{equation}
where $Y_{T}$ and $Y_{S}$ are the teacher output and the student output, respectively, $|\cdot|$ denotes the sum of all elements in a matrix, $*$ denotes element-wise multiplication, and $\epsilon$ is a small constant to avoid division by zero. The Dice loss ranges from 0 to 1, where lower values indicate higher similarity.

By minimizing the Dice loss, the student model learns to mimic the teacher model's output and thus segment \gls{sar} images without requiring annotated data. The Dice loss provides a soft and smooth supervision signal for the student model, as it considers true positives in the numerator and true positives, false positives, and false negatives in the denominator. This Dice loss formula helps solve the issue of imbalanced training data and does not require defining weighting parameters between different classes (in our case, the ground and water). Besides, the function works well for binary segmentation tasks \citep{soomro2018strided}.

\subsection{The Backpropagation Algorithm}
The backpropagation algorithm is the algorithm that updates the student weights based on the Dice loss. The backpropagation algorithm consists of two steps: forward and backward propagation. In the forward propagation, we compute the teacher output, the student output, and the Dice loss for a given pair of optical and \gls{sar} images. In the backward propagation, we calculate the gradient of the Dice loss with respect to the student weights and update the weights using an optimizer.

The steps of the backpropagation algorithm are as follows:

\begin{itemize}
    \item Step 1: Given a pair of optical and \gls{sar} images ($X_O$, $X_S$), we feed $X_O$ to the teacher model and $X_S$ to the student model. We then obtain $Y_T$ and $Y_S$ as the teacher's and student's outputs, respectively.
    \item Step 2: Compute $\mathcal{L}_{Dice}$ using $Y_T$ and $Y_S$ as inputs.
    \item Step 3: Compute $\partial\mathcal{L}_{Dice}/\partial W_S$ using the chain rule, where $W_S$ are the student weights.
    \item Step 4: Update $W_S$ using an optimizer (e.g., Adam \citep{kingma2014adam}).
    \item Step 5: Repeat steps 1-4 for all pairs of optical and \gls{sar} images in training set until convergence.
\end{itemize}

%% file: 5_evaluation.tex
\section{Evaluation}\label{sec:evaluation}

We evaluated the performance of \deepaqua using \acrshort{sar}-Vertical-Horizontal (VH) imagery downloaded from Google Earth Engine \citep{gorelick2017google}.

\subsection{Training, Validation, and Testing Datasets}

\begin{figure}[H]
    \centering
    \includegraphics[width=\linewidth]{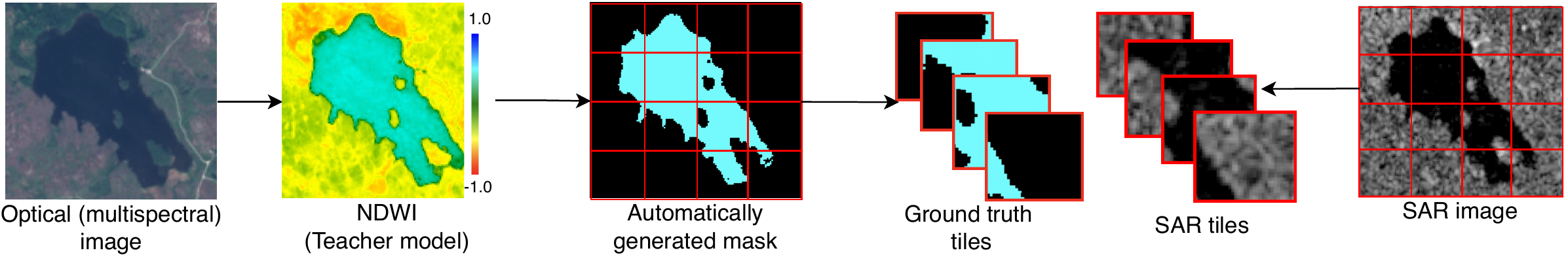}
    \caption{The process of automatically generating the training set. We create multispectral-\acrshort{sar} pairs by splitting the satellite images into smaller tiles.}
    \label{fig:dataset}
\end{figure}

To train \deepaqua, we used Sentinel-1 (\acrshort{sar}) and Sentinel-2 (multispectral) images of the entire county of \"Orebro in Sweden. \"Orebro has an area of $(\sim 8550$ km$^{2})$. We used Sentinel-2 multispectral images and then applied \gls{ndwi} to generate water masks. To this end, we first split the entire \"Orebro region into tiles of $64 \times 64$ pixels. Each pixel had a resolution of $10$ meters. Then, we repeated the same procedure to generate a \gls{sar} dataset using Sentinel-1 images from the same region. This resulted in a total of $45\,500$ multispectral-\acrshort{sar} pairs. Figure \ref{fig:dataset} illustrates how we generated the data to train our model. Once we generated all the data, we randomly selected 80\% of the tiles to create a training set, and we took the remaining 20\% to create a validation set.

\begin{figure}[htbp]
    \centering
    \includegraphics[width=\linewidth]{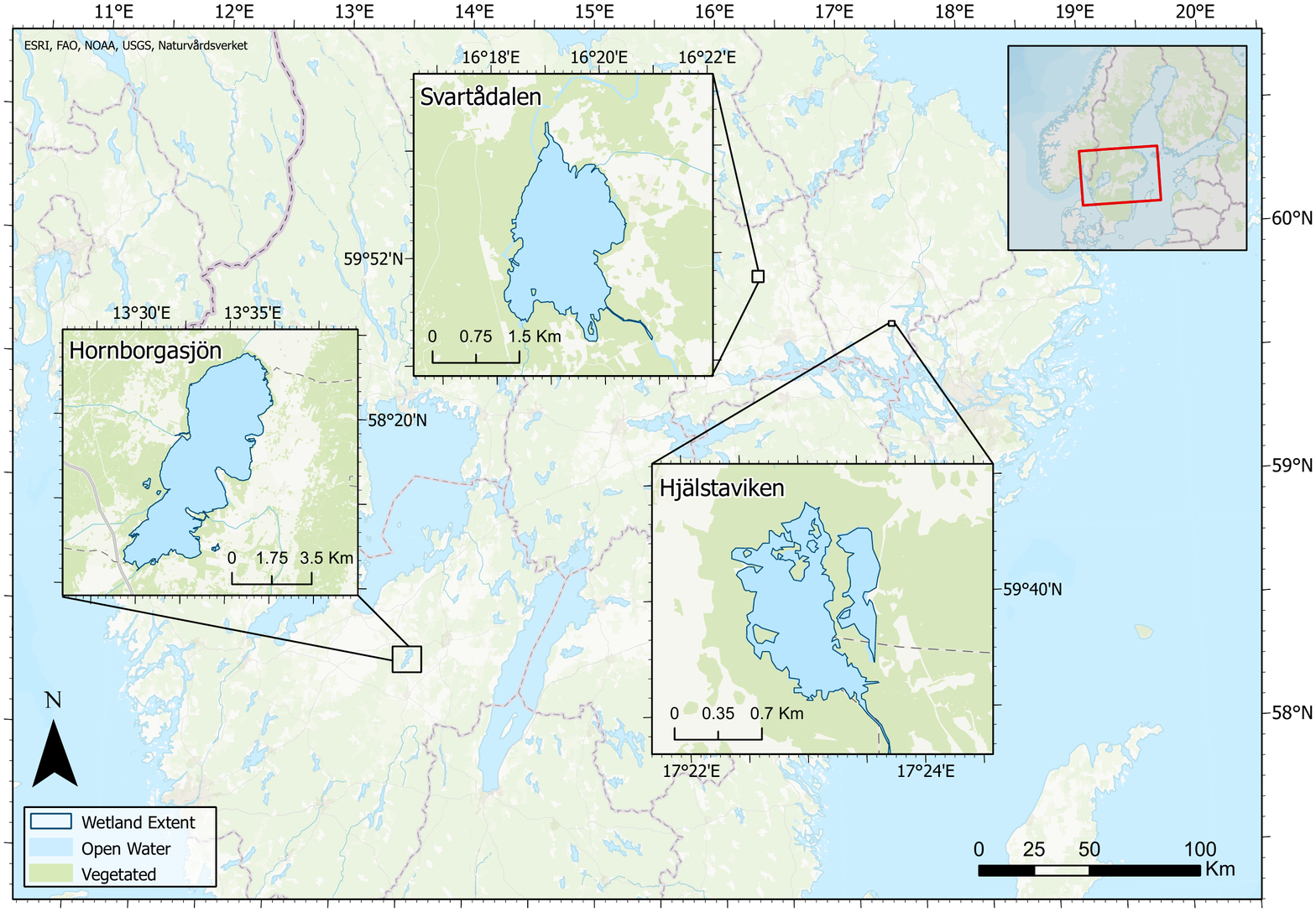}
    \caption{Map of Sweden containing the location of the three wetlands with manually annotated data that compose the testing set.}
    \label{fig:wetlands_map}
\end{figure}

Our testing set is composed of imagery from Svart{\aa}dalen, Hj\"alstaviken and Hornborgasj\"on, three wetlands that belong to the Ramsar convention, as shown in Figure \ref{fig:wetlands_map}. The three wetlands are located in flat areas of Southern and Western Sweden. These are shallow wetlands that are not connected to any important river stream and are rather fed by small streams with low flow rates. The wetlands are convered by emerging grassy vegetation mainly consisting of mires, and open water bodies. The grassy vegetation is often flooded during the rainy season and during spring when the wetlands receive snowmelt from upstream. Some borders of the wetlands have also tree canopies that do not allow the penetration of C-band signals. However, these areas are not connected to the grassy areas and are far from the open water bodies. Svart{\aa}dalen is a mixed wetland complex of $1\,977$ ha comprised of mires, bogs, and fens. Hj\"alstaviken is a limnic complex of $808$ ha. Hornborgasj\"on is a human-made mire complex of $6\,197$ ha, one of the largest single nature conservation projects ever carried out in Sweden \citep{gunnarsson2014swedish,matthews1993ramsar}. We manually delineated the water in radar imagery from the study sites. Each site contains 40 images between 2018 and 2022. We excluded the months of January, February, March, and December to avoid images that contained snow.

\subsection{Evaluation Metrics}

To quantitatively evaluate the performance of our semantic segmentation model on radar imagery of wetlands, we employed a set of evaluation metrics as detailed in \citep{everingham2015pascal}. These metrics, derived from \gls{tp}, \gls{tn}, \gls{fp}, and \gls{fn} counts, measure both pixel-level accuracy and the overall quality of the segmented regions.

\begin{enumerate}
    \item \textbf{\acrfull{pa}}: This metric computes the proportion of correctly classified pixels in the entire image. It provides an overall sense of how well the model is performing but may not capture errors distributed across different classes effectively: $\frac{\text{TP} + \text{TN}}{\text{TP} + \text{TN} + \text{FP} + \text{FN}}$.
    
    \item \textbf{\acrfull{iou}}: Also known as the Jaccard index, \gls{iou} offers a measure of the overlap between the predicted and true areas, with values ranging from 0 (no overlap) to 1 (perfect overlap): $\frac{\text{TP}}{\text{TP} + \text{FP} + \text{FN}}$
    
    \item \textbf{Precision}: This quantifies the proportion of positive identifications (i.e., water pixels) that were actually correct. A high precision indicates that the model has fewer false positives: $\frac{\text{TP}}{\text{TP} + \text{FP}}$
    
    \item \textbf{Recall}: Also known as sensitivity, recall measures the proportion of actual positives (water pixels in ground truth) that are identified correctly. A high recall means the model has fewer false negatives: $\frac{\text{TP}}{\text{TP} + \text{FN}}$
    
    \item \textbf{F1-Score}: The harmonic mean of precision and recall, the F1-score gives a balanced measure of the model's performance, especially when the class distribution is imbalanced: $2 \times \frac{\text{Precision} \times \text{Recall}}{\text{Precision} + \text{Recall}}$
\end{enumerate}

By employing these metrics, we aim to comprehensively evaluate our model's capabilities, considering both the finer details and the broader context of water segmentation in radar imagery.

\subsection{Baseline Method}

We compared the performance of \deepaqua against Otsu's method \citep{otsu1979threshold}. We selected this method because it is unsupervised, aligning with our study's central theme of working with non-manually annotated data, making it distinct from the other methods discussed in Section \ref{sec:related_work}.

We decided to concentrate on unsupervised methods because \deepaqua's core advantage is its capacity to train without manual annotations. Introducing supervised methods into our comparison would present complications related to the quality of training data, which might shift attention away from our primary goal: demonstrating the effectiveness of a strong unsupervised solution.

Otsu's method is a technique for automatically determining the optimal threshold value for image segmentation or binarization. This method and other thresholding approaches, such as the one described in \citep{carvalho2011new}, seek to find a threshold value that augments the distinction between an image's foreground (water, in our context) and background (soil). Otsu's method computes the variance between these two classes of pixels for every conceivable threshold value. The threshold that mitigates the variance within each class while amplifying the variance between the classes is adopted as the prime choice. We executed our experiments using OpenCV's Python implementation of the Otsu method.

Radar images are naturally noisy, so filtering techniques are often used to improve the segmentation process \citep{tan2023self,zhou2020application,li2020improved}. We incorporated a Gaussian filter variation of the Otsu method in response to these recognized challenges. This variation assists in diminishing the noise impact, ensuring more accurate segmentation without drastically diverging from the raw data. It is noteworthy that while the Gaussian filter aids in reducing noise, our core objective remains to accentuate the self-supervised potential of our proposed CNN model. This focus on self-supervision aligns with our choice of a minimal preprocessing strategy.

We benchmarked \deepaqua using the Dynamic World dataset \citep{brown2022dynamic}, which classifies land cover based on Sentinel-2 optical imagery. We identified \textit{water} and \textit{flooded vegetation} as positive classes, with others as negative. We favored Dynamic World for its five-day update frequency, in contrast to the yearly updates of datasets like Esri \citep{karra2021global} and the European Space Agency \citep{zanaga2022esa}. However, its reliance on optical imagery could limit vegetated water detection compared to radar-based methods.

\subsection{Implementation Details}

We implemented our methods using PyTorch \citep{paszke2019pytorch} with an Adam \citep{kingma2014adam} optimizer with a learning rate of $5 \times 10^{-5}$. We minimized the Dice loss function \citep{dice1945measures} as described in Section \ref{sec:loss}. We trained our method for 20 epochs with a batch size of 32 on a MacBook Pro with an M1 processor and 16GB of RAM. The total training time was 277 minutes.

We trained the \gls{cnn} using the raw data to show the true power of our approach in handling complex and noisy images. We did not apply any filters or preprocessing techniques to clean the original \gls{sar} images, except for removing outlier pixel values by discarding values lower than percentile $1$ and higher than percentile $99$. We also applied min-max scaling to bring the pixel values to the range $[0, 1]$. We could increase prediction performance using techniques such as image denoising, data augmentation, image contrasting, transfer learning, and model ensembling; however, in this paper, we focused only on the potential of the \gls{ndwi} water masks to train \gls{sar}-based \glspl{cnn}.

We selected and tuned the hyperparameters of our method using a grid search based on the performance of the validation set. For the learning rate, we tried the values [$1 \times 10^{-6}, 5 \times 10^{-6} 1 \times 10^{-5}, 5 \times 10^{-5}, 5 \times 10^{-4}, 1 \times 10^{-3}, 5 \times 10^{-3}$], and for the batch size, we tried the values [$1, 2, 4, 8, 16, 32, 64, 128, 256$]. We stopped at $20$ epochs because the model converged at this time. We also experimented with other water indexes such as \gls{mndwi} \citep{xu2006modification}, \gls{awei} \citep{feyisa2014automated} and \gls{hrwi} \citep{yao2015high}. We applied the Otsu method with a Gaussian filter using a $5\times 5$ kernel size.

One of our challenges was that the models trained on 2018 data only showed good performance for 2018 and 2019 but poor performance in 2020, 2021, and 2022. For this reason, we had to train two models based on data from different years. The first model was trained on satellite images taken on July 4th, 2018, and worked well for all 2018-2019 images. The second model was trained on satellite images taken on June 23rd, 2020, working well for all 2020-2022 images. However, after inspecting the images, we realized that the \gls{sar} images from 2018-2019 had more speckle and noise than those from 2020-2022, possibly due to an adjustment on the Sentinel-1 sensors. Therefore, we provide a pre-trained version of our model for both 2018 and 2020.

The code, testing dataset, and pre-trained models are available at\\ {\tt https://github.com/melqkiades/deep-wetlands}.

\subsection{Quantitative Results}

As Table \ref{tab:quantitative_results} shows, \deepaqua outperforms the Otsu models on accuracy, \gls{iou}, recall, and F1-score by a significant margin on all three study areas, demonstrating the effectiveness of our approach in leveraging cross-modal knowledge distillation without requiring annotated data.

\begin{table}[H] %
\begin{center}
\scalebox{0.54}{
\begin{tabular}{l r r r r r r r r r r r r r r r r }
  \toprule
  \textbf{Model} & \multicolumn{5}{c}{\textbf{Svart{\aa}dalen}} & \multicolumn{5}{c}{\textbf{Hj\"alstaviken}} & \multicolumn{4}{c}{\textbf{Hornborgasj\"on}}  \\
  \cmidrule(lr){2-6} \cmidrule(lr){7-11} \cmidrule(lr){12-16}
   & \textbf{\acrshort{pa}} & \textbf{\acrshort{iou}}    & \textbf{Prec}  & \textbf{Recall}  & \textbf{F1} & \textbf{\acrshort{pa}} & \textbf{\acrshort{iou}}    & \textbf{Prec}  & \textbf{Recall}  & \textbf{F1} & \textbf{\acrshort{pa}} & \textbf{\acrshort{iou}}    & \textbf{Prec}  & \textbf{Recall}  & \textbf{F1} \\
\midrule
Dynamic World          & 0.92 & 0.59 & 0.92 & 0.63 & 0.75 & 0.90 & 0.35 & 0.57 & 0.47 & 0.51 & 0.88 & 0.57 & 0.91 & 0.60 & 0.73 \\
Otsu & 0.90            & 0.64 & 0.69 & 0.89 & 0.78 & 0.81 & 0.34 & 0.36 & 0.85 & 0.51 & 0.79 & 0.49 & 0.65 & 0.66 & 0.66 \\
Otsu + Gaussian filter & 0.93 & 0.73 & 0.75 & \textbf{0.96} & 0.84 & 0.83 & 0.38 & 0.40 & \textbf{0.89} & 0.55 & 0.83 & 0.56 & 0.73 & 0.70 & 0.72 \\
\midrule
\deepaqua-NDWI         & \textbf{0.97} & \textbf{0.88} & \textbf{0.98} & 0.90 & \textbf{0.93} & \textbf{0.96} & 0.68 & 0.81 & 0.81 & \textbf{0.81} & \textbf{0.98} & \textbf{0.94} & \textbf{0.98} & 0.96 & \textbf{0.97} \\
\deepaqua-MNDWI        & \textbf{0.97} & 0.85 & 0.95 & 0.89 & 0.92 & 0.95 & 0.68 & 0.78 & 0.84 & \textbf{0.81} & \textbf{0.98} & 0.93 & 0.96 & \textbf{0.97} & 0.96 \\
\deepaqua-AWEI         & \textbf{0.97} & 0.84 & \textbf{0.98} & 0.85 & 0.91 & \textbf{0.96} & 0.68 & \textbf{0.86} & 0.77 & \textbf{0.81} & \textbf{0.98} & \textbf{0.94} & \textbf{0.98} & 0.95 & \textbf{0.97} \\
\deepaqua-HRWI         & \textbf{0.97} & 0.86 & 0.97 & 0.88 & 0.92 & \textbf{0.96} & \textbf{0.69} & 0.82 & 0.81 & \textbf{0.81} & \textbf{0.98} & \textbf{0.94} & 0.97 & 0.96 & \textbf{0.97} \\
       
\bottomrule
\end{tabular}
}
\caption{Semantic segmentation performance over various wetland areas in Sweden. The best performance is highlighted in bold.\label{tab:quantitative_results}}
\end{center}
\end{table}

Using data from Table \ref{tab:quantitative_results}, which delineates results for the three study areas, the metrics presented here are derived from a weighted aggregation of \deepaqua's performance, factoring in the varying sizes of each site. The \deepaqua-NDWI model, which is the best overall performer, demonstrates an approximate accuracy of 98\%, reflecting its consistent efficacy across different terrains. An \gls{iou} of 92\% accentuates its precision in mapping the overlap between predicted and actual water extents. With a precision of 97\%, the model robustly pinpoints true water pixels in its positive predictions, and a recall of 94\% indicates its capacity to identify most water pixels within the dataset. This balance between precision and recall results in an F1 score of 96\%. While the performance of the various \deepaqua models is similar, we can see that \deepaqua-NDWI has a slight edge. This could be due to the \gls{ndwi} index it uses, which has a 10-meter resolution, compared to the 20-meter resolution of \gls{mndwi} and \gls{awei}.

In comparison, using the same weighted aggregation approach, the Otsu model with a Gaussian filter achieves an accuracy of 92\%, an \gls{iou} of 72\%, and an F1-score of 84\%. Its precision is marked at 74\%. However, it exhibits a high recall of 96\%, indicating its proficiency in recognizing water bodies, albeit with some propensity to misclassify non-water areas. While the Dynamic World model struggles to detect vegetated water, its precision surpasses the Otsu model across all three study areas, indicating that its water pixel predictions are often accurate.

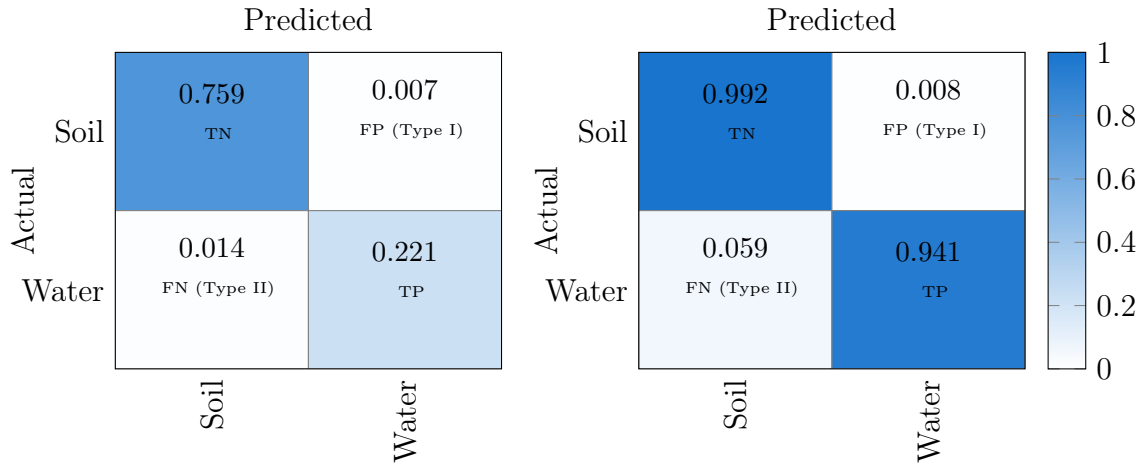
\begin{figure}[H]
    \centering

    \begin{subfigure}[b]{0.49\textwidth}
        \centering
        \begin{tikzpicture}
        \begin{axis}[
            width=\textwidth,
            colormap={bluewhite}{color=(white) rgb255=(24,116,205)},
            xlabel=Predicted,
            ylabel=Actual,
            xticklabels={Soil, Water}, 
            xtick={0,...,1}, 
            xtick style={draw=none},
            yticklabels={Soil, Water}, 
            ytick={0,...,1}, 
            ytick style={draw=none},
            enlargelimits=false,
            xticklabel style={
                rotate=90
            },
            nodes near coords={
                \pgfmathprintnumber[fixed,precision=3]\pgfplotspointmeta{} \\ \tiny \coordinateLabel{} %
            },
            nodes near coords style={
                yshift=-7pt,
                font=\small, 
                align=center %
            },
            point meta min=0, %
            point meta max=1, %
            xlabel style={at={(axis description cs:0.5,1.3)},anchor=north},
        ]

        \newcommand{\coordinateLabel}[1]{
            \pgfmathtruncatemacro\labelindex{\coordindex}
            \ifcase\labelindex
                TN %
            \or
                FP (Type I) %
            \or
                FN (Type II) %
            \or
                TP %
            \fi
        }

        \addplot[
            matrix plot,
            mesh/cols=2, 
            point meta=explicit,draw=gray
        ] table [meta=C] {
            x y C
            0 0 0.759
            1 0 0.007
            0 1 0.014
            1 1 0.221
        };
        \end{axis}
    \end{tikzpicture}
        \caption{Matrix normalized by all pixel count.}
        \label{fig:confusion_matrix_all}
    \end{subfigure}
    \hfill
    \begin{subfigure}[b]{0.49\textwidth}
        \centering
        \begin{tikzpicture}
        \begin{axis}[
            width=\textwidth,
            colormap={bluewhite}{color=(white) rgb255=(24,116,205)},
            xlabel=Predicted,
            ylabel=Actual,
            xticklabels={Soil, Water}, 
            xtick={0,...,1}, 
            xtick style={draw=none},
            yticklabels={Soil, Water}, 
            ytick={0,...,1}, 
            ytick style={draw=none},
            enlargelimits=false,
            colorbar,
            xticklabel style={
                rotate=90
            },
            nodes near coords={
                \pgfmathprintnumber[fixed,precision=3]\pgfplotspointmeta{} \\ \tiny \coordinateLabel{} %
            },
            nodes near coords style={
                yshift=-7pt,
                font=\small, 
                align=center %
            },
            point meta min=0, %
            point meta max=1, %
            xlabel style={at={(axis description cs:0.5,1.3)},anchor=north},
        ]

        \newcommand{\coordinateLabel}[1]{
            \pgfmathtruncatemacro\labelindex{\coordindex}
            \ifcase\labelindex
                TN %
            \or
                FP (Type I) %
            \or
                FN (Type II) %
            \or
                TP %
            \fi
        }

        \addplot[
            matrix plot,
            mesh/cols=2, 
            point meta=explicit,draw=gray
        ] table [meta=C] {
            x y C
            0 0 0.992
            1 0 0.008
            0 1 0.059
            1 1 0.941
        };
        \end{axis}
    \end{tikzpicture}
        \caption{Matrix normalized by actual class pixel count.}
        \label{fig:confusion_matrix_class}
    \end{subfigure}

    \caption{Comparative confusion matrices for \deepaqua's performance. Subfigure \ref{fig:confusion_matrix_all} displays normalization against the total pixel count, while Subfigure \ref{fig:confusion_matrix_class} emphasizes performance metrics specific to each class.}
    \label{fig:two_confusion_matrices_sub}
\end{figure}

\deepaqua demonstrates effective water surface detection and consequent water surface extent estimation, with errors being a small portion of its results, as evident from the confusion matrices in Figure \ref{fig:two_confusion_matrices_sub}. The 0.7\% \gls{fp} rate indicates a controlled overestimation of aquatic regions. On the other hand, the 1.4\% \gls{fn} rate suggests some missed water bodies. These figures highlight the model's ability to distinguish between soil and water, especially when considering the challenges of detecting both open and vegetated water surfaces in dynamic wetland environments. The results not only point to areas for further refinement but also validate \deepaqua's utility in the field of satellite-based water monitoring.

\subsection{Qualitative Results}

\begin{figure}[H]
    \centering
    \includegraphics[width=\linewidth]{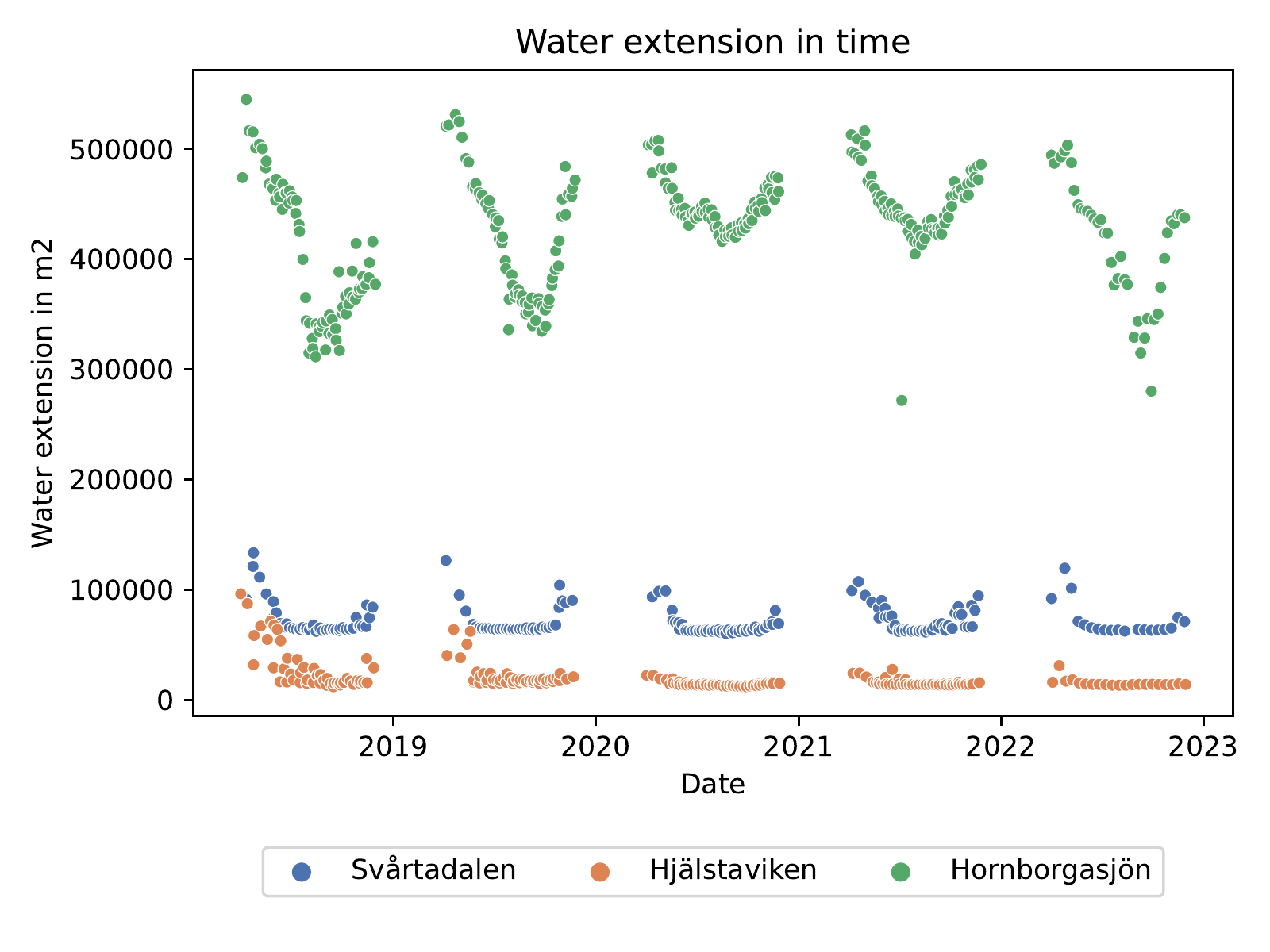}
    \caption{Dots indicate the water extension of three Swedish wetlands over 2018-2022. Only the months of April through November are considered.}
    \label{fig:water_extension_flacksjon}
\end{figure}

We applied \deepaqua to assess the surface water extent in three study areas over multiple dates from 2018 to 2022. Figure \ref{fig:water_extension_flacksjon} shows the water extent in the wetlands as measured by our model, revealing an annual cycle. Typically, the wetlands experience increased water levels during spring due to snowmelt, which diminishes as summer arrives. As autumn approaches, the wetland surface water extent increases once again.

To underscore the accuracy of \deepaqua, Figure \ref{fig:image_predictions} offers predictions for the three study areas across different months. The leftmost column displays the input \gls{sar} image, while the rightmost presents manually annotated ground truth. Intermediate columns feature predictions from our model variations, particularly on \deepaqua-\acrshort{ndwi} due to its superior performance.

The top row captures the Sv{\aa}rtadalen wetland during summer in July 4th, 2018. The Otsu method's difficulty with speckle noise is evident, though filtering mitigates some of this. The \deepaqua model appears resilient to this noise and aligns closely with the ground truth. The middle row captures the Hjälstaviken wetland during autumn on October 4th, 2020. This image is more noisy than the previous one, causing challenges for the Otsu method. Instead of reducing the noise, the Gaussian filter amplifies it. \deepaqua deals well with the noise, offering clear predictions. However, as indicated by the red pixels, some areas at the top of the image missed water detection. The bottom row shows the Hornborgasj\"on wetland in spring on April 19th, 2021. Both Otsu methods falter due to noise, while \deepaqua largely avoids it, though some \gls{fp} emerge, as indicated by the cyan pixels.

Notably, \deepaqua accurately predicts surface water extent and identifies ``land islands'' within the wetlands. While it is not flawless, with some \gls{fp} and \gls{fn} errors evident, its noise reduction capability is commendable and achieved without resorting to additional filtering or preprocessing. As the red and cyan pixels in Figure \ref{fig:image_predictions}, errors often appear around wetland shores or where water levels are low. In these regions, differentiating between water and soil in \gls{sar} images can be challenging due to the mixed signals from the water-soil interface. We emphasize that \deepaqua does not use filtering or pre-processing techniques on the \gls{sar} images.

\addtolength{\tabcolsep}{-4pt}    
\newcommand{\addpic}[1]{\includegraphics[width=6.1em]{#1}}
\newcolumntype{C}{>{\centering\arraybackslash}m{6.1em}}
\begin{figure} %
\begin{tabular}{l*5{C}@{}}
& \acrshort{sar} image & Otsu  & Otsu + filter & \deepaqua & Ground truth  \\ 
A & \addpic{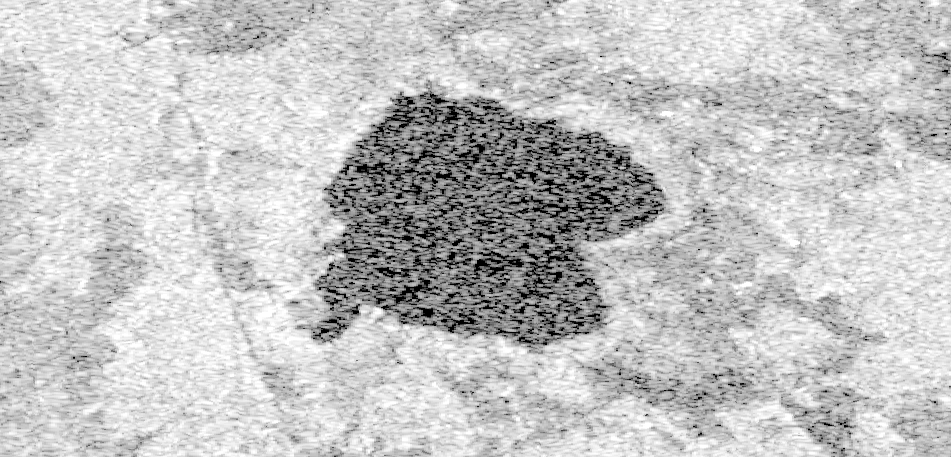} & \addpic{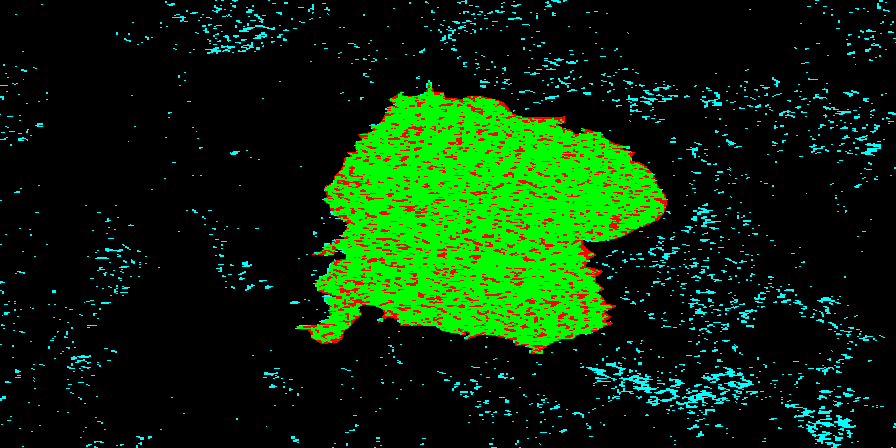} & \addpic{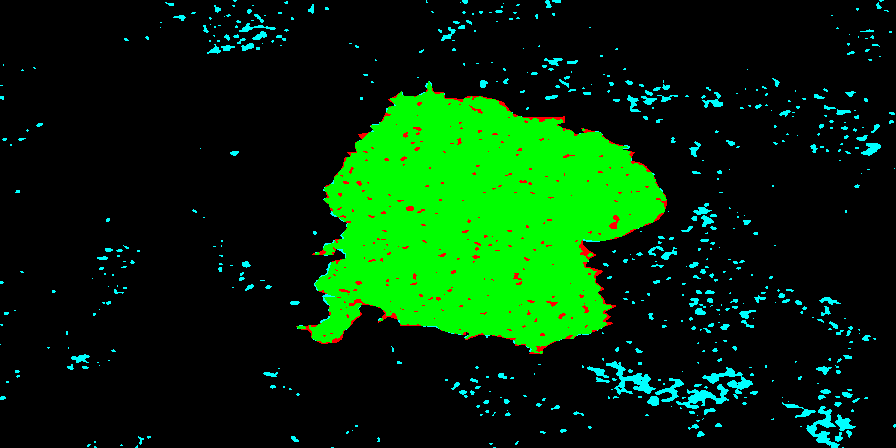} & \addpic{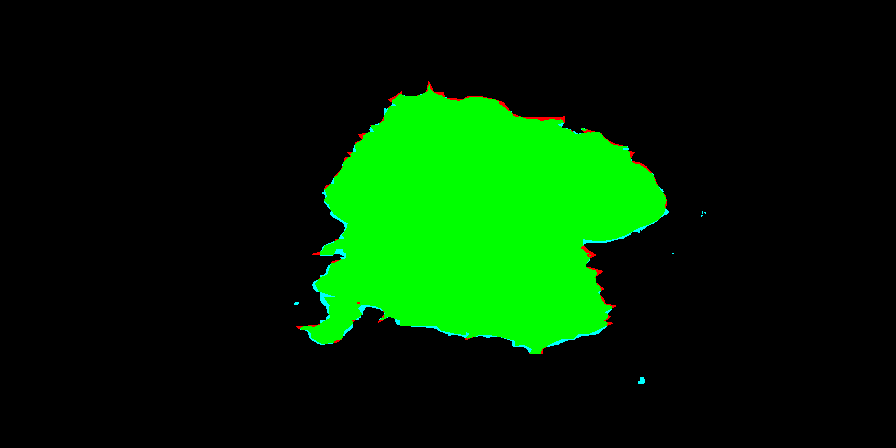} & \addpic{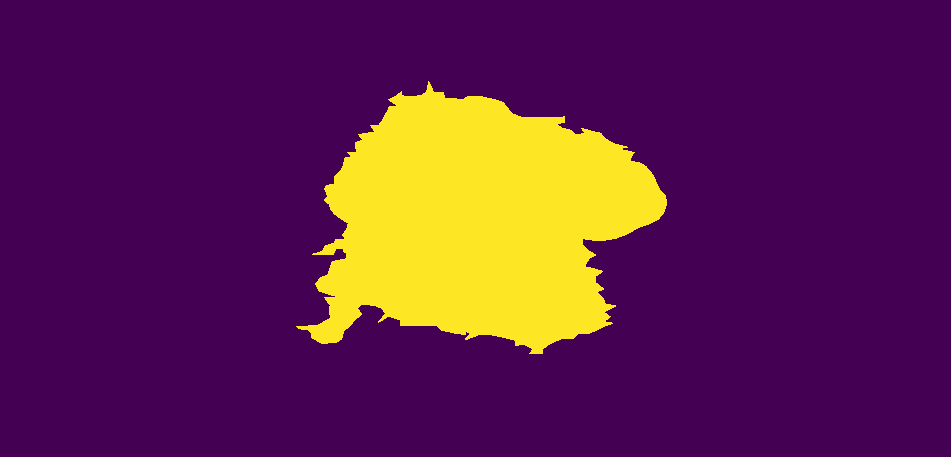} \\
B & \addpic{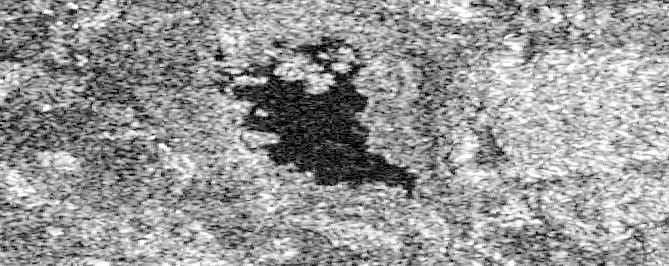} & \addpic{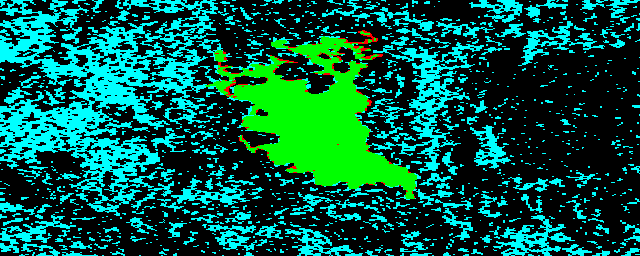} & \addpic{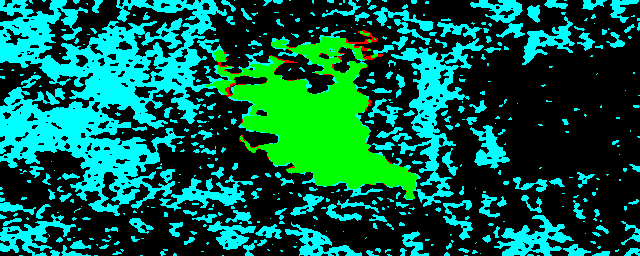} & \addpic{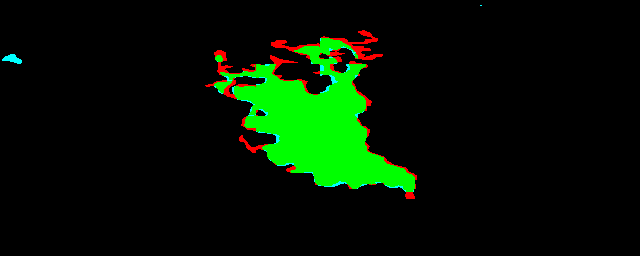} & \addpic{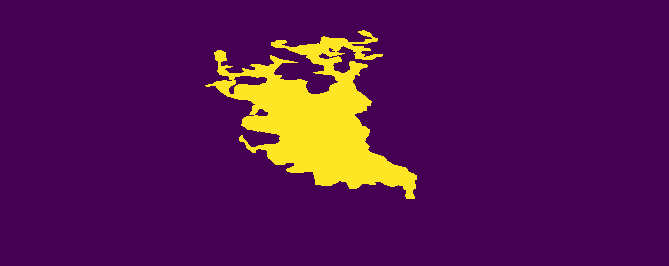} \\ 
C & \addpic{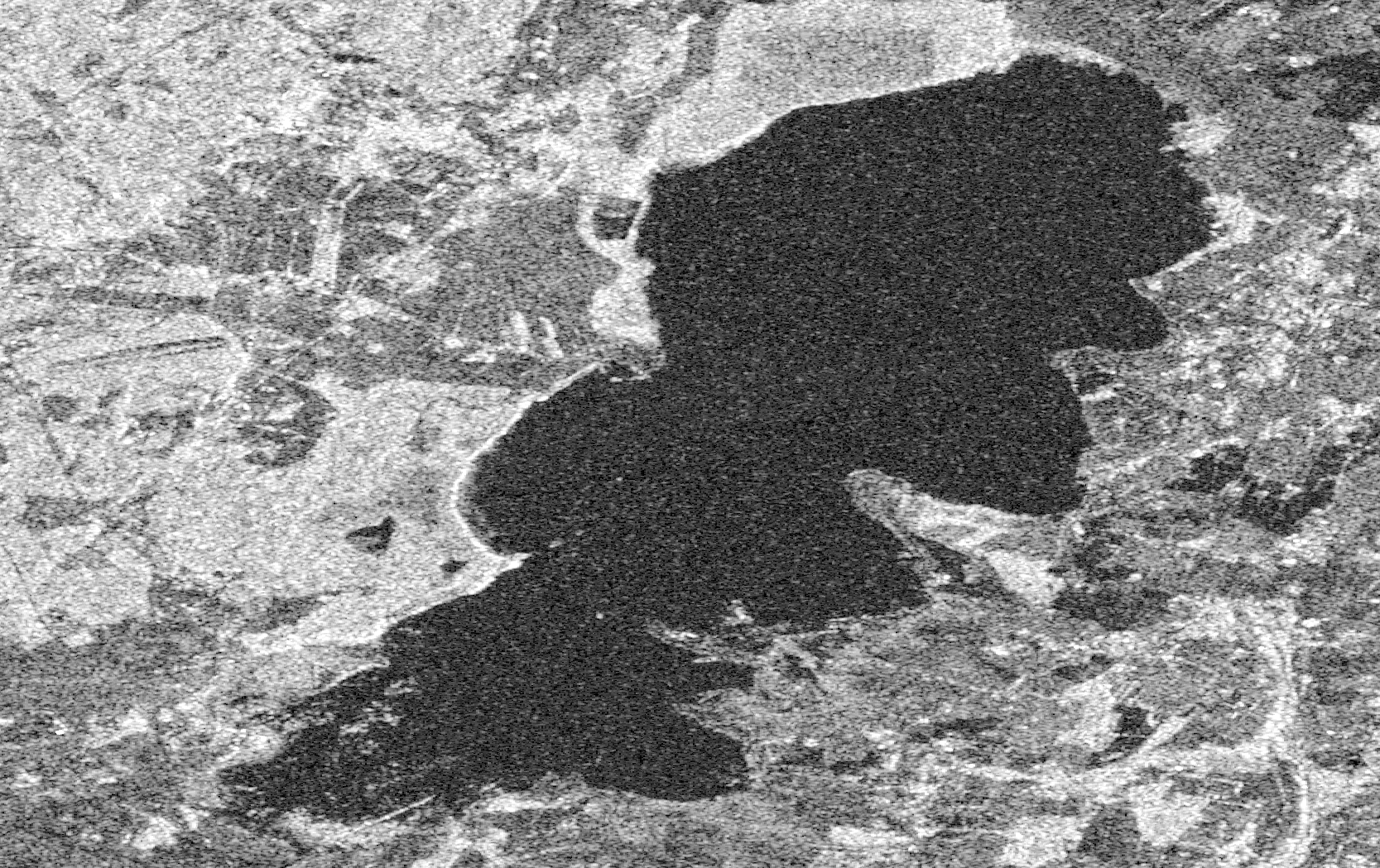} & \addpic{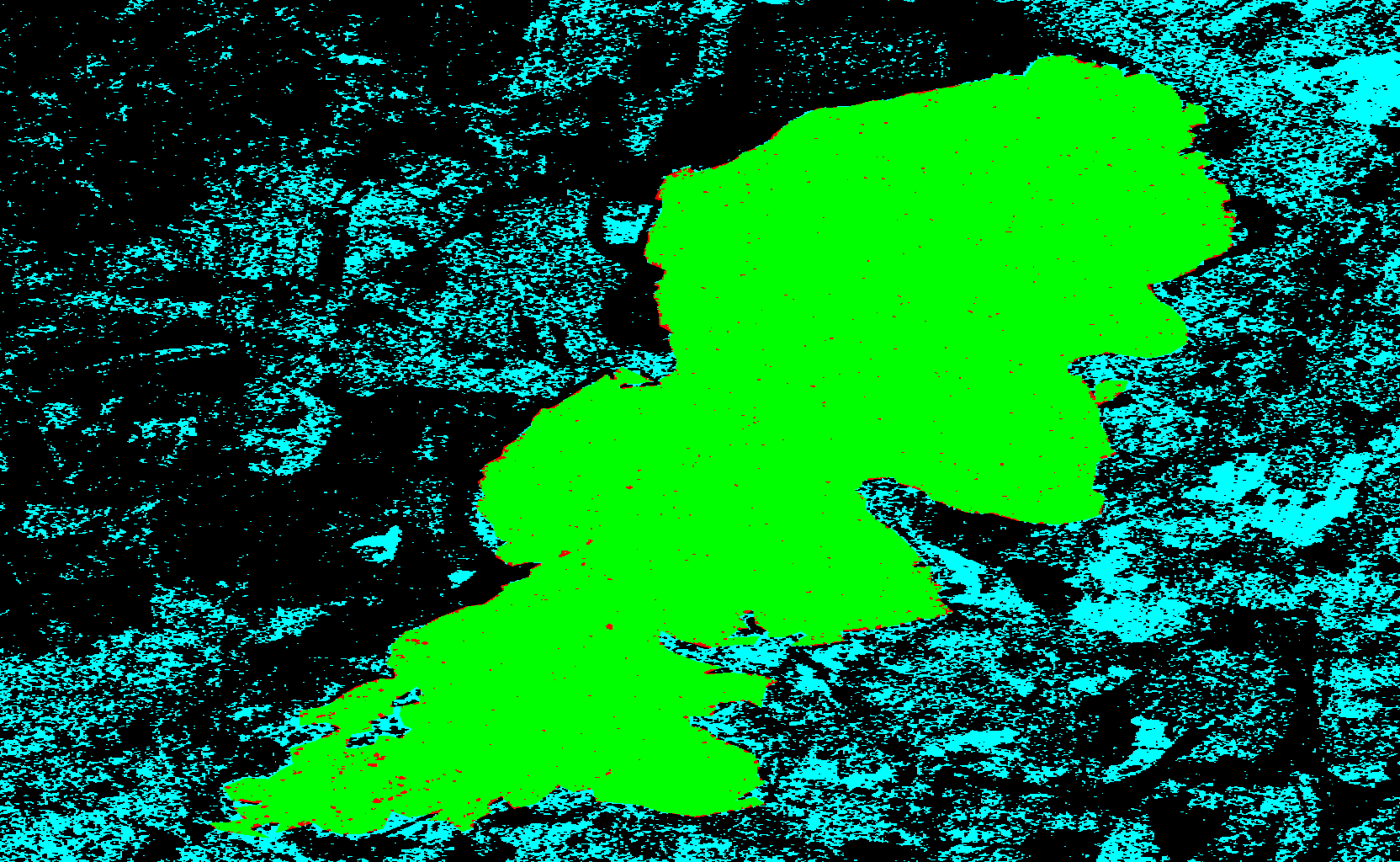} & \addpic{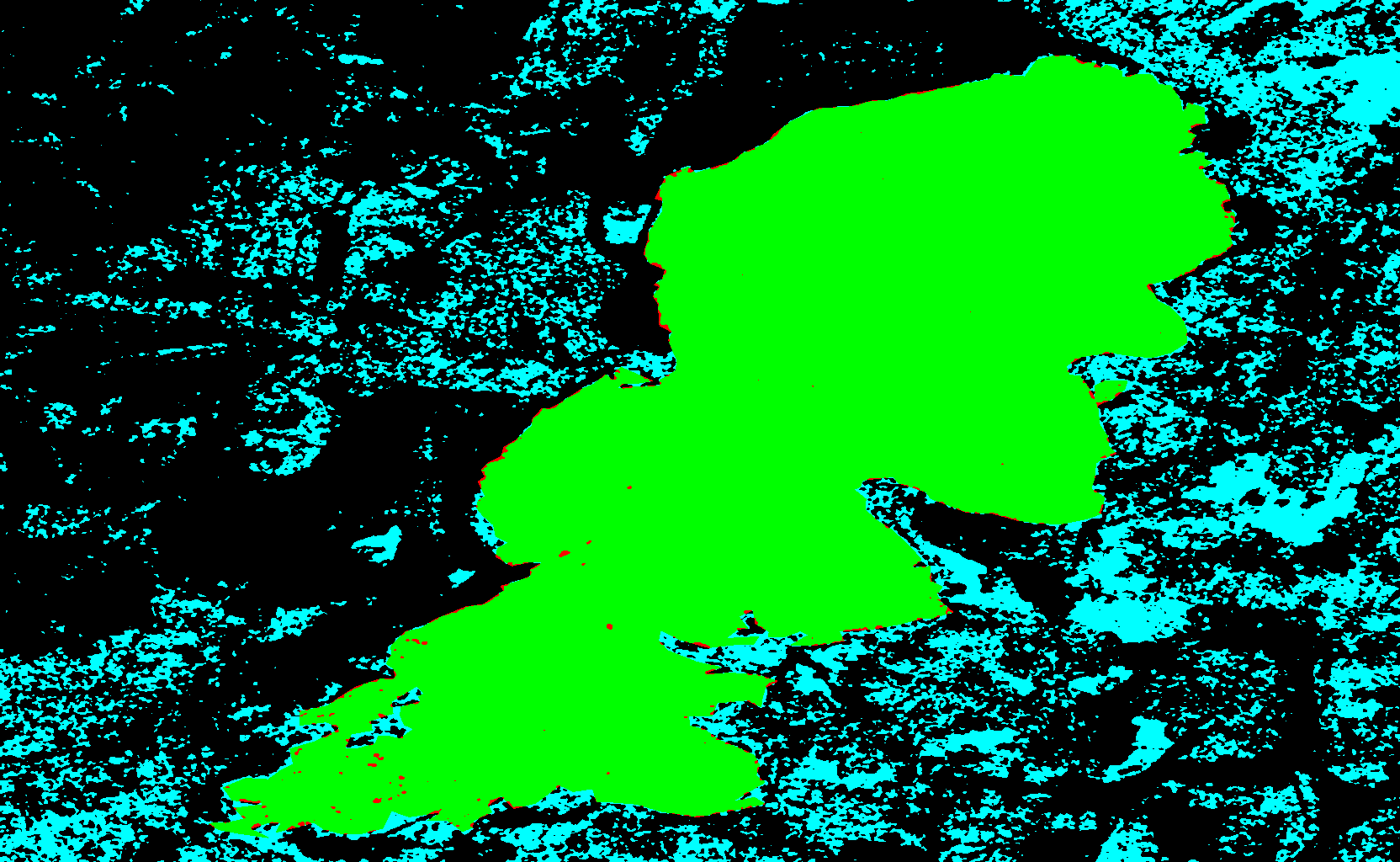} & \addpic{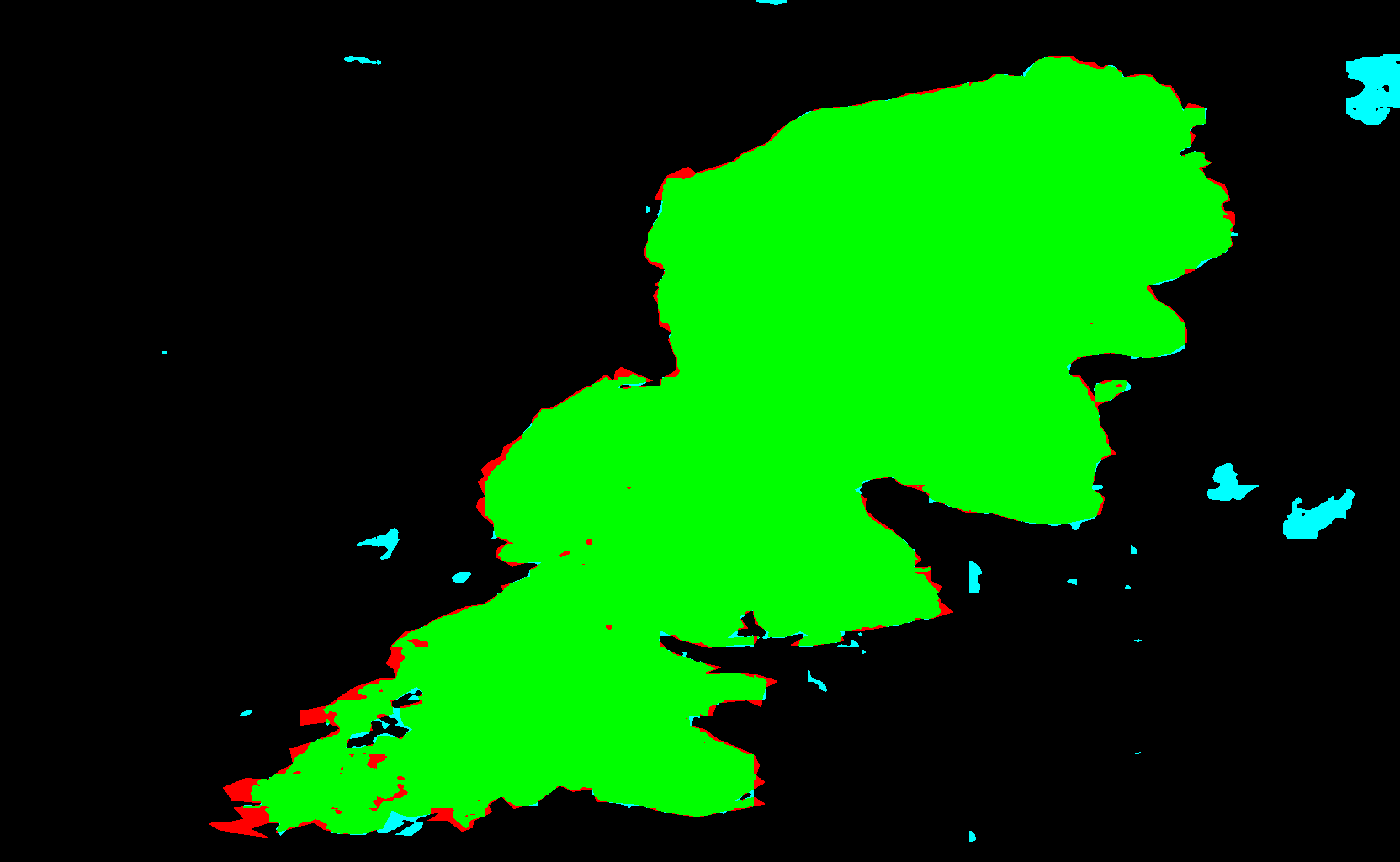} & \addpic{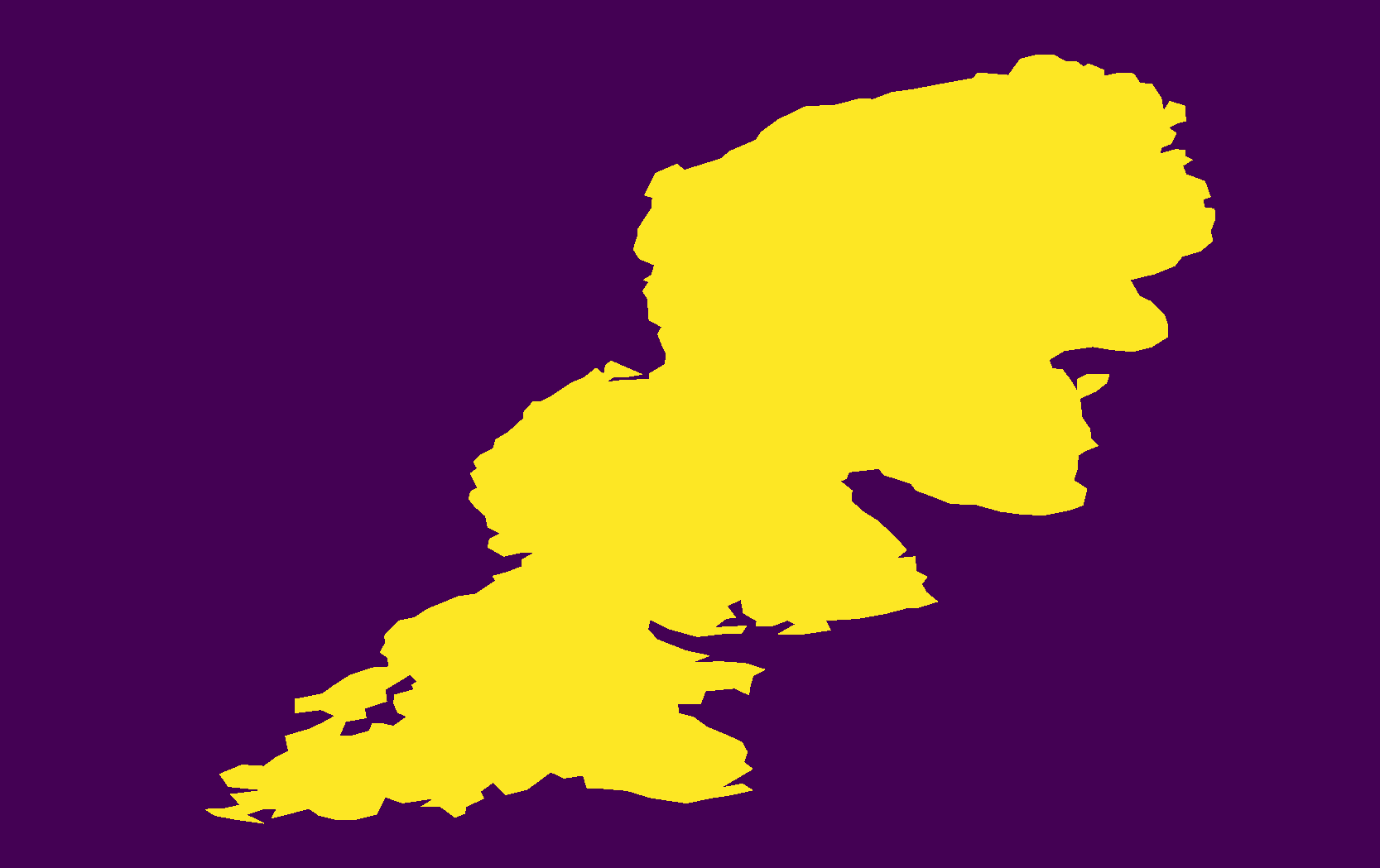} \\ 
\end{tabular}
\caption{Illustration of the performance of our model for different areas and times of the year: Sv{\aa}rtadalen, July 4th, 2018 (A); Hj\"alstaviken, October 4th, 2020; and Hornborgasj\"on April 19th, 2021 (C). From left to right: original \acrshort{sar} images, segmentation using Otsu's method, segmentation using Otsu's method with a Gaussian filter, segmentation using \deepaqua, and manually annotated data. Green pixels denote \gls{tp}, cyan pixels denote \gls{fp}, red pixels denote \gls{fn} and black pixels denote \gls{tn}.}
\label{fig:image_predictions}
\end{figure}
\addtolength{\tabcolsep}{4pt}

To summarize, \deepaqua excels in recognizing water signatures from \gls{sar} images. While the method effectively tracks wetland transitions over time, it does not inherently distinguish between open and vegetated waters due to \gls{sar}'s imaging nature. Although not the primary focus, coupling our approach with water index methods like \gls{ndwi} can differentiate these water types. However, we recommend exploring radar sensors with more extended wavelengths than the C-band for dense vegetation like mangroves.

%% file: 6_conclusion.tex
\section{Conclusion}\label{sec:conclusions}

We present \deepaqua, which is a novel method that uses cross-modal knowledge distillation to train a \gls{cnn} for semantic segmentation of water in \gls{sar} imagery without requiring annotated data. Our method consists of two models: a teacher model that creates \gls{ndwi} water masks from optical images and a student model that learns to segment water in \gls{sar} images. We used U-Net to implement the student model. The teacher and student models are trained jointly by minimizing the Dice loss between their outputs. Our experiments confirmed that our model can accurately segment images, confidently detecting water. The model here is trained and tested in a marsh wetland environment. However, it can be applied to wetlands with emerging vegetation allowing the penetration of C-Band \gls{sar} signals. Future studies may adapt the approach for radar sensors with longer wavelengths to expand the applicability in wetlands with larger vegetation, such as Mangroves. The use of this model can aid applications where water detection is crucial, such as flooding detection, river and lake mapping, and water availability assessments in time and space.

%% file: acknowledgements.tex
\section*{Acknowledgements}

This work was supported by Digital Futures. We gratefully acknowledge the Bolin Centre for Climate Research for providing us with a GPU server and the National Academic Infrastructure for Super computing in Sweden for giving us access to their high computing cluster. We thank Ezio Cristofoli, Johanna Hansen, and Ioannis Iakovidis for their ideas and for annotating data.